\documentclass[10pt, conference, compsocconf]{IEEEtran}
\usepackage[letterpaper, left=1in, right=1in, top=0.75in, bottom=1in]{geometry}
\usepackage{cite}
\usepackage{amsmath,amssymb,amsfonts}
\usepackage{algorithmic}
\usepackage{graphicx}
\usepackage{textcomp}
\usepackage{xcolor}
\def\BibTeX{{\rm B\kern-.05em{\sc i\kern-.025em b}\kern-.08em
    T\kern-.1667em\lower.7ex\hbox{E}\kern-.125emX}}
\usepackage{xurl}

\usepackage{multirow}

\usepackage{colortbl,array,xcolor}

\usepackage{threeparttable}
\usepackage{dblfloatfix}    

\newcommand{\figref}[1]{Fig.~\ref{fig:#1}}
\newcommand{\tabref}[1]{Table~\ref{tab:#1}}

\newcommand{\etal}{\textit{et al.}}
\usepackage{hyperref} 

\begin{document}

\title{Differentially Private Cross-camera Person Re-identification}

\author{
    \IEEEauthorblockN{
        Lucas Maris\IEEEauthorrefmark{1},
        Yuki Matsuda\IEEEauthorrefmark{1}\IEEEauthorrefmark{2},
        Keiichi Yasumoto\IEEEauthorrefmark{1}\IEEEauthorrefmark{2}
    }
        
    \IEEEauthorblockA{\IEEEauthorrefmark{1}
        Nara Institute of Science and Technology, Nara, Japan
    }
    \IEEEauthorblockA{\IEEEauthorrefmark{2}
        RIKEN Center for Advanced Intelligence Project AIP, Tokyo, Japan\\
        Email: \{lucas.maris.lo3, yukimat, yasumoto\}@is.naist.jp
    }
}

\maketitle

\begin{abstract}
Camera-based person re-identification is a heavily privacy-invading task by design, benefiting from rich visual data to match together person representations across different cameras. This high-dimensional data can then easily be used for other, perhaps less desirable, applications. We here investigate the possibility of protecting such image data against uses outside of the intended re-identification task, and introduce a differential privacy mechanism leveraging both pixelisation and colour quantisation for this purpose. We show its ability to distort images in such a way that adverse task performances are significantly reduced, while retaining high re-identification performances.

\end{abstract}


\section{Introduction}
Cities keep growing, creating logistic challenges in terms of traffic management, city planning, and tourism. In parallel, the interest in smart cities, i.e., cities that use information and communication technologies to transform the city and its governance, and achieve a positive impact on the community inhabiting it~\cite{deakin2011smartcity}, is peaking, with both large and small-scale projects being developed around the globe.

Japan, for instance, sees its population decrease and its demography evolve, with a decrease of the relative working-age population and a general movement toward metropolitan areas~\cite{schieder2021japan}. These changes and the resulting shift in citizen expectations make it highly valuable to dispose of comprehensive data regarding human flows within the city. This data can then be exploited by, e.g., policymakers for urban planning decisions~\cite{schrotter2020urbanplanning}, transport or tourism companies for commercial ends~\cite{khan2021transport}, or individuals for route planning purposes~\cite{isoda2020routeplanning}.

One of the most commonly available sensors in cities is a video camera. Often installed primarily as a means of passive surveillance, with the aim of disposing of a record of events for understanding past incidents, their feeds have the potential to be leveraged for real-time traffic or crowd monitoring.

In this context, cross-camera person re-identification has been extensively studied over the past decade, as the computer vision task of matching individuals from different camera perspectives. Notwithstanding promising applications in security, planning, or tourism areas, little concern has been paid to the glaring privacy concerns this raises, and thus the social acceptance of such systems. This study aims to build a privacy-aware cross-camera re-identification system.

By empirically comparing the effect of different types and magnitudes of noise, including both traditional image obfuscation methods and a new differential privacy method, on the inputs to a state-of-the-art re-identification model, we evaluate the feasibility of conducting the paradoxical task of re-identifying people without effectively invading their privacy. We illustrate this goal by means of an additional empirical evaluation of the ability to predict demographic attributes, such as gender, age, or ethnicity, from these very same noised inputs. 


The purpose is to distort visual data in such a way that it remains maximally useful for the specific intent of video camera controllers, here chosen to be cross-camera re-identification, while making it minimally useful for other purposes, here defined as common demographic classification tasks. The ability to control the privacy leakage of re-identification systems is key to their social acceptance, which requires building trustful relationships with citizens, customers, or individuals at large.

Our contributions are as follows: (1) we formulate a strict image differential privacy mechanism leveraging both pixelisation and colour quantisation; (2) we highlight the robustness of centroid-based re-identification models against noise; and (3) we show that the use of our image differential privacy mechanism allows for nearly state-of-the-art re-identification performances while significantly reducing the utility of images for adverse tasks.


\section{Related Work}
\label{sec:relatedwork}
\subsection{Person re-identification (reID)}
The recognition of individuals across different visual snapshots from different points of view has become a traditional computer vision task over the last 20 years~\cite{ye2022reidreview}. While still a matter of designing informative appearance signatures for individuals~\cite{bazzani2010appearance, farenzena2010appearance}, the advent of deep convolutional neural networks~\cite{krizhevsky2012cnn} has shifted this problem to yet another learning problem~\cite{yi2014deepreid}.

Notwithstanding its many variations, the task is essentially defined as the retrieval of all occurrences of a given individual within a set of videos, often captured from different angles, with only a single snapshot of that individual to work from. This is usually formalised as an image-retrieval task, where the aim is to train a model on a \textit{training} set such that it can rank images from the \textit{gallery} set in order of their similarity to a given image from the \textit{query} set.

As the input data is usually a set of videos, this reID definition glances over the object detection, tracking and segmentation aspects to focus more heavily on how to transform images into effective vector representations and on how to score such representations against one another. Recent state-of-the-art reID systems achieve maximum performances on traditional reID datasets such as Market1501~\cite{zheng2015market1501} or MSMT17~\cite{wei2017msmt17} by answering these questions with CNN feature extractors~\cite{he2016resnet} and triplet loss~\cite{hermans2017tripletloss}, combined with various other techniques, e.g., the use of spatial-temporal data ~\cite{wang2018spatialtemporal}, large-scale pre-training~\cite{fu2021unsupervised} or attention mechanisms~\cite{chen2022attention}.


\subsection{Differential privacy (DP)}
Over the last decade, differential privacy~\cite{dwork2014dp} has become the single most popular way of modelling formal data privacy, due to its ability to provide quantifiable protection against arbitrary risks. By perturbing computations over statistical databases, it promises indistinguishability between databases and plausible deniability to every individual composing such databases, providing them with roughly the same privacy that would result from having their data removed from the database.

Recent work regarding differential privacy has aimed at extending its desirable properties to other forms of data, e.g., location~\cite{andres2013lgeodp} or deep neural network models~\cite{abadi2016nndp}. With unstructured data making up most of today's data landscape, a line of work regarding image differential privacy has also emerged, with studies leveraging pixelization~\cite{fan2018pixeldp}, autoencoders~\cite{liu2021autoencoderdp} or generative adversarial networks~\cite{li2021gandp}. As noted in a recent survey regarding differential privacy for unstructured data~\cite{zhao2022unstructureddp}, the common approach is to vectorize unstructured data into a structured form, which can then be obfuscated with conventional DP methods.

\section{Method}
In this paper, we consider the inherent privacy threat that originates from storing image-representations of individuals for tracking across multiple video feeds. We expect this to be useful for applications where the aim is to collect itinerary data from pedestrians in such a way that the specifics of every individual are protected; for instance, one may wish to know the percentage of people that visit point B shortly after point A, without needing to know whether this percentage includes a specific individual that could be recognised from their visual representation. 

In other words, we consider the threat of an attacker gaining access to a collection of images from individuals, which were collected for the purpose of re-identification. To lessen the gravity of such a breach, we aim to distort collected images such that they remain maximally useful for the cross-camera reID task but minimally useful for other tasks. The adverse tasks we focus on in this study include gender, age, and ethnicity identification.

\subsection{DP-obfuscation}
To provide a quantifiable privacy guarantee on the gallery images, we extend the pixel-level differential privacy definition first introduced as DP-Pix~\cite{fan2018pixeldp}. As it provides privacy directly on pixel values, it has since been deemed too strict of a definition, destroying too much of the data's utility in the privacy process~\cite{liu2021autoencoderdp, li2021gandp, zhao2022unstructureddp}. This being precisely our goal here, we further restrict its definition; instead of providing indistinguishability between same-sized images differing by at most $m$ pixels, we here aim for indistinguishability between same-sized images differing in \emph{any} amount of pixels. \newline

\textbf{Definition 1.} \emph{$\varepsilon$-Image DP}: a randomized mechanism $\mathcal{M}$ gives $\varepsilon$-image differential privacy if for any two images $i$ and $j$ of same dimension, and for any possible output $R \subseteq \text{Range}(\mathcal{M})$,
\begin{equation}
    \text{Pr}[\mathcal{M}(i) \in R] \leq \text{exp}(\varepsilon) \ \text{Pr}[\mathcal{M}(j) \in R]
\end{equation}

As shown in \cite{dwork2014dp}, the Laplace mechanism can achieve such a guarantee, provided $\mathcal{M}$ is defined as the noisy function:
\begin{equation}
    \mathcal{M}(x) = f(x) + n \newline \text{, where } n \sim Laplace\left(0, \frac{\Delta f}{\varepsilon}\right)
\end{equation}

The exact amount of noise $n$ is to be calibrated to the sensitivity $\Delta f$ of function $f$. We generalise and extend the DP-Pix definition of this function $f$; instead of defining $f$ as the pixelisation of grayscale image $x$, we define $f$ as the identity function applied to RGB image $x$, with optional pixelisation and colour quantisation parameters $b$ and $c$. The sensitivity of this function is then:
\begin{equation}
    \Delta f = \frac{wh}{b^2} \left( \frac{256}{c}-1 \right)^3
\end{equation}

Where $w$ and $h$ are the width and height of images, $b$ defines the amount of pixelisation to be applied to images, and $c$ characterises the amount of colours to be kept in images. These additional dimensionality reduction parameters are introduced to reduce the magnitude of the sensitivity, increased by our broader neighborhood definition. If $b=1$ and $c=1$, the sensitivity is equivalent to that of the identity function.

This privacy mechanism is to be applied directly on stored images, prior to their use for re-identification. By protecting the image data upfront with a quantifiable privacy guarantee, their sensitivity can be mitigated. Even in the event of a data leak, the utility of training, query, and gallery samples is expected to be decreased for other applications.

\subsection{Centroid-reID}
Despite the higher noise-level introduced by our strict pixel-level differential privacy definition, we wish to be able to link obfuscated image representations with their corresponding identity. One recently introduced method to improve on re-identification and image-retrieval tasks at large is to average training samples into mean centroid representations, i.e., aggregated class representations. By shifting the task from ranking specific identity-instances to classifying into actual identities, which arguably also makes more sense in practical applications, state-of-the-art re-identification performances can be achieved on classic reID datasets~\cite{wieczorek2021centroids}.

The training of a centroid-based re-identification model relies on the Centroid Triplet Loss function, which is formulated below. It aims to minimise the distance between embedding $f(A)$ of a training sample $A$ and the class centroid $c_P$ for the class that sample belongs to, while maximising the distance between said embedding $f(A)$ and class centroid $c_N$ for a class the sample does not belong to.
\begin{equation}
    \mathcal{L}_{CTL} = \left[||f(A)-c_P||^2_2 - ||f(A)-c_N||^2_2 + \alpha_c\right]_+
\end{equation}

Class centroids $c_k$ are simply defined as the mean embedding of all samples available at a given point; during training, these are the items of same class as the currently considered sample within a mini-batch; during testing, these are all gallery samples for a given class. If we denote the available class samples as $\mathcal{S}_k$, the class centroid $c_k$ is:
\begin{equation}
    c_k = \frac{1}{|\mathcal{S}_k|} \sum_{x_i \in S_k} f(x_i)
\end{equation}

We here postulate that by using averaged individual representations, the re-identification system can be made more robust against introduced noise. We expect such aggregated representations to be able to magnify the identity-specific latent features that remain underneath the noise. To this end, we train centroid-based models directly onto noised training images, and test them with noised sets of gallery and query samples.

\subsection{Adverse tasks}
To evaluate the privacy protection offered by our system beyond the theoretical privacy budget $\varepsilon$, we additionally consider adverse tasks in the form of demographic attribute classification, specifically gender, age range, and ethnic group. As such tasks are both common and feasible with very limited inputs, we believe them a good way to evaluate the practical privacy protection offered by our mechanism.

Attribute classification is implemented as a simple fully connected layer on top of a ResNet50v2~\cite{he2016resnet} backbone, itself pretrained on ImageNet~\cite{russakovsky2015imagenet}.

\section{Results}
\label{sec:results}
In this section, we use an $\varepsilon$-Image DP-protected dataset for both our target task, person re-identification, and adversary tasks, here chosen to be gender, age, and ethnicity classification. We show how little our target task suffers from the privacy protection applied to the dataset, while the performance of adversary tasks drops more drastically.

We evaluate both the reID and gender classification performances after obfuscation on Market1501~\cite{zheng2015market1501}, a common reID dataset, whose balanced attribute labels are unfortunately limited to unbalanced gender annotations~\cite{lin2019marketattr}. To further evaluate the degree of obfuscation provided by our method, we additionally evaluate the performance of gender, age, and ethnicity classification models after applying the same obfuscation to FairFace~\cite{karkkainen2021fairface}, a recent face image dataset that provides balanced gender, age, and ethnicity annotations.

\subsection{ReID}
\begin{figure*}[h]
\centering
\includegraphics[width=\linewidth]{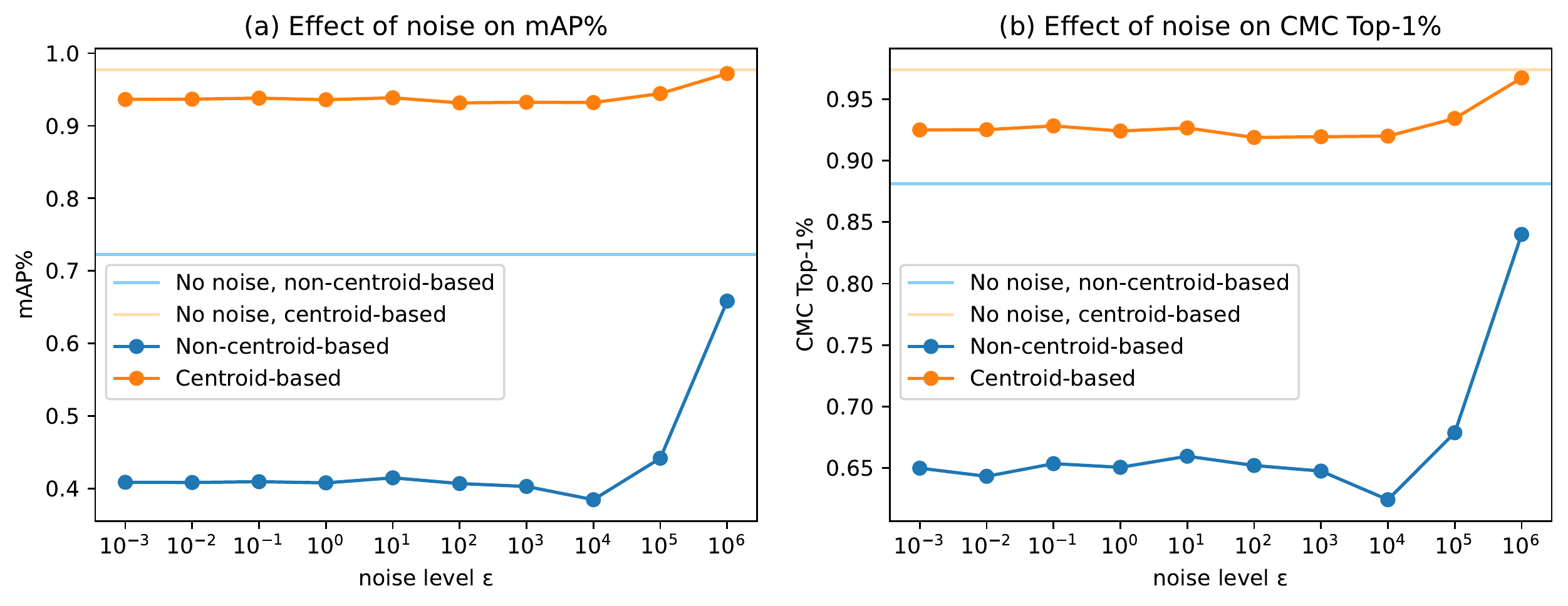}
\caption{Market1501 reID results, with b=1, c=64, $\Delta f$=221184.}
\label{fig:reid_b1c64}
\end{figure*}

To properly get a feel for the effect of pixelisation and quantisation parameters $b$ and $c$, we consider 3 parameter combinations; $b=1$ and $c=64$, $b=2$ and $c=32$, and $b=4$ and $c=16$, which yield sensitivity values $\Delta f$ of similar magnitude. Noise levels $\varepsilon$ were made to vary between $10^{-3}$ and $10^6$. \figref{reid_b1c64} shows the performance of both a regular and a centroid-based reID model on data noised with the first set of parameters. \tabref{reid_b2c32} and \tabref{reid_b4c16} summarise the results for the second and third sets of parameters, as these yield visually similar graphs.



\begin{table}[h]
\centering
\caption{Market1501 reID results, with b=2, c=32, $\Delta f$=702464}
\label{tab:reid_b2c32}
\begin{tabular}{c|cc|cc}
\hline
& \multicolumn{2}{c|}{\textbf{Regular}} & \multicolumn{2}{c}{\textbf{Centroid-based}} \\
\hline
\textbf{Noise $\varepsilon$} & \textbf{mAP\%} & \textbf{Top-1\%} & \textbf{mAP\%} & \textbf{Top-1\%} \\
\hline
$10^{-3}$ & 19.7\% & 39.0\% & 81.6\% & 78.6\% \\
$1$ & 19.4\% & 38.2\% & 81.0\% & 77.6\% \\
$10^3$ & 19.7\% & 38.4\% & 81.6\% & 78.4\% \\
$10^6$ & 43.9\% & 68.2\% & 94.3\% & 93.2\% \\
\hline
none & 76.8\% & 90.7\% & 98.1\% & 97.7\% \\
\hline
\end{tabular}
\end{table}

\begin{table}[h]
\centering
\caption{Market1501 reID results, with b=4, c=16, $\Delta f$=1728000}
\label{tab:reid_b4c16}
\begin{tabular}{c|cc|cc}
\hline
& \multicolumn{2}{c|}{\textbf{Regular}} & \multicolumn{2}{c}{\textbf{Centroid-based}} \\
\hline
\textbf{Noise $\varepsilon$} & \textbf{mAP\%} & \textbf{Top-1\%} & \textbf{mAP\%} & \textbf{Top-1\%} \\
\hline
$10^{-3}$ & 5.0\% & 11.1\% & 41.5\% & 34.8\% \\
$1$ & 4.1\% & 8.9\% & 35.7\% & 29.2\% \\
$10^3$ & 4.0\% & 8.7\% & 35.1\% & 28.7\% \\
$10^6$ & 11.7\% & 24.8\% & 67.6\% & 62.2\% \\
\hline
none & 66.1\% & 83.9\% & 97.0\% & 96.4\% \\
\hline
\end{tabular}
\end{table}

It is striking from our experiments that centroid-based reID models offer high robustness to noise. While non-centroid-based models suffer, on average, a 43.4\%, a 74.5\%, and a 93.6\% mAP decrease, for each of our parameter combinations in ascending $b$ order, centroid-based models only suffer an average 4.2\%, 17.0\% and 62.4\% mAP decrease, respectively. These averages are computed on mAP metrics obtained with  $\varepsilon$ varying between $10^{-3}$ and $10^{3}$. As \figref{reid_b1c64} illustrates, this robustness allows to conserve a mean mAP of 93.6\% even at noise levels $\varepsilon\leq10^{3}$.

It is also quite apparent from experiments that there exists a hard limit on the effect of pixel-based noising, with $\varepsilon$ values lower than $10^{3}$ yielding more or less the same results as those observed at the $\varepsilon=10^{3}$ level. Considering the fact that our sensitivities $\Delta f$ are themselves of the order of $10^{5}$ or $10^{6}$, which naturally increases the relative scale of $\varepsilon$ values, and that differentially private noise is meant to be scaled to these values, this is in line with the \textit{``All Small Epsilons Are Alike''} observation by Dwork \etal~\cite{dwork2014dp}.

\subsection{Attribute classification}
\subsubsection{Market1501}
Using the same noised Market1501 images to perform gender classification yields the results summarised in \tabref{market_attr}.



\begin{table}[h]
\centering
\caption{Market1501 gender classification accuracy}
\label{tab:market_attr}
\begin{tabular}{c|c|c|c}
\hline
\multirow{2}{*}{\textbf{Noise $\varepsilon$}} & \multicolumn{3}{c}{\textbf{Accuracy\%}} \\
\cline{2-4}
& \textit{b=1, c=64} & \textit{b=2, c=32} & \textit{b=4, c=16} \\
\hline
$\mathit{10^{-3}}$ & 65.9\% & 63.5\% & 61.6\% \\
$\mathit{1}$ & 65.9\% & 63.4\% & 61.8\% \\
$\mathit{10^3}$ & 65.8\% & 64.0\% & 61.4\% \\
$\mathit{10^6}$ & 73.1\% & 65.5\% & 62.9\% \\
\hline
\textit{none} & 77.2\% & 78.3\% & 74.8\% \\
\hline
\end{tabular}
\end{table}

The initial 81.3\% gender classification accuracy, obtained on the original Market1501 dataset, gets affected both by the dimensionality reduction parameters b and c, as shown in the unnoised row of the table, and the additional noise added on top of this transformation. As with our reID results, the effect of added noise stagnates when setting $\varepsilon \leq 10^{3}$. The average accuracy decrease, computed with $\varepsilon$ varying between $10^{-3}$ and $10^{3}$, for each parameter combination in ascending b order, is 14.6\%, 19.0\%, and 18.0\%. Considering that 57.0\% of the total samples are men, which implies a trivial classifier would achieve a 57.0\% accuracy, we can see our privacy mechanism brings gender prediction performances on Market1501 sensibly closer to chance-level.

\subsubsection{Fairface}
Likewise, we apply the same transformations to the images from the Fairface dataset, on which we then train gender, age, and ethnicity classification models. As the images from this dataset are significantly larger, being 224x224 against 64x128 for Market1501, we intentionally use a higher pixelisation level b, to make the sensitivity $\Delta f$ of the images from these different datasets more comparable. For brevity's sake, we only consider a single parameter combination $b=4$ and $c=32$, which yields sensitivity $\Delta f=1075648$. \tabref{fairface_attr} goes over the observed results.

\begin{table}[h]
\centering
\caption{Fairface classification accuracy, \\ with b=4 c=32, $\Delta f$=1075648}
\label{tab:fairface_attr}
\begin{tabular}{c|c|c|c}
\hline
\multirow{2}{*}{\textbf{Noise $\varepsilon$}} & \multicolumn{3}{c}{\textbf{Accuracy\%}} \\
\cline{2-4}
& \textit{Gender} & \textit{Age} & \textit{Ethnicity} \\
\hline
$\mathit{10^{-3}}$ & 66.0\% & 31.6\% & 31.8\% \\
$\mathit{1}$ & 66.3\% & 32.0\% & 31.8\% \\
$\mathit{10^3}$ & 65.7\% & 31.8\% & 31.9\% \\
$\mathit{10^6}$ & 67.0\% & 32.8\% & 32.5\% \\
\hline
\textit{none} & 76.9\% & 39.7\% & 42.4\% \\
\hline
\textit{chance} & 50.0\% & 11.1\% & 14.3\% \\
\hline
\end{tabular}
\end{table}

Again, the initial classification accuracies of 79.8\%, 43.3\% and 46.1\%, obtained on the original Fairface dataset for each task respectively, get affected by both the dimensionality reduction step as well as the noising step. The effect of noise also tends to stabilise with $\varepsilon \leq 10^{3}$; the average accuracy decreases, when $\varepsilon$ varies between $10^{-3}$ and $10^{3}$, are 14.1\%, 19.5\% and 25.7\%, for each task respectively.

\section{Discussion}
\label{sec:discussion}
\subsection{Effect of parameters $b$ and $c$}
This section further studies the individual effects of the pixelisation parameter $b$ and quantisation parameter $c$, in order to get a better understanding of the performance changes that are to be attributed to our privacy mechanism $\varepsilon$-Image DP, and those that are to be attributed to more traditional computer vision transformations.

\begin{figure*}[h]
\centering
\includegraphics[width=\linewidth]{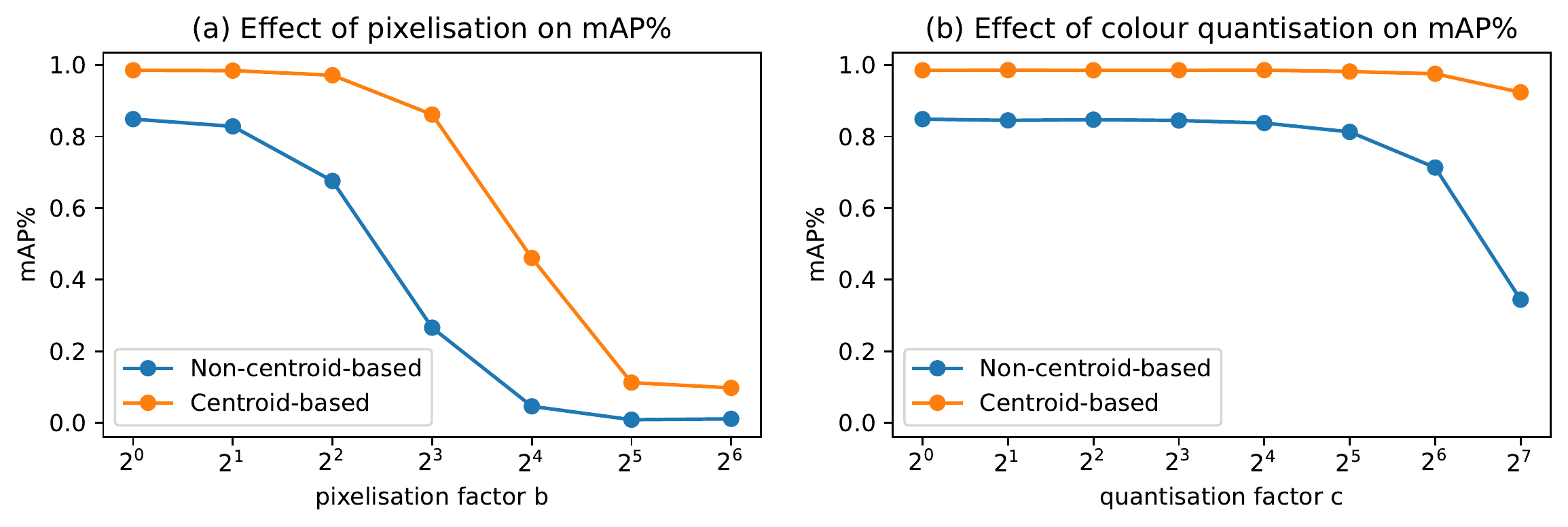}
\caption{Market1501 reID results, with varying $b$ and $c$.}
\label{fig:market_reid_dimreduc}
\end{figure*}

\figref{market_reid_dimreduc} shows the effect of increasingly large values for $b$ and $c$ on reID performances for Market1501 images. While heavy pixelisation destroys the utility of images for reID, with mAP dipping under 50\% when $b \geq 2^4$, heavy colour quantisation has nearly no impact on performances, even at $c = 2^7$. In both cases, a centroid-based reID model proves more effective at identifying people than a regular reID model.

\begin{figure*}[h]
\centering
\includegraphics[width=\linewidth]{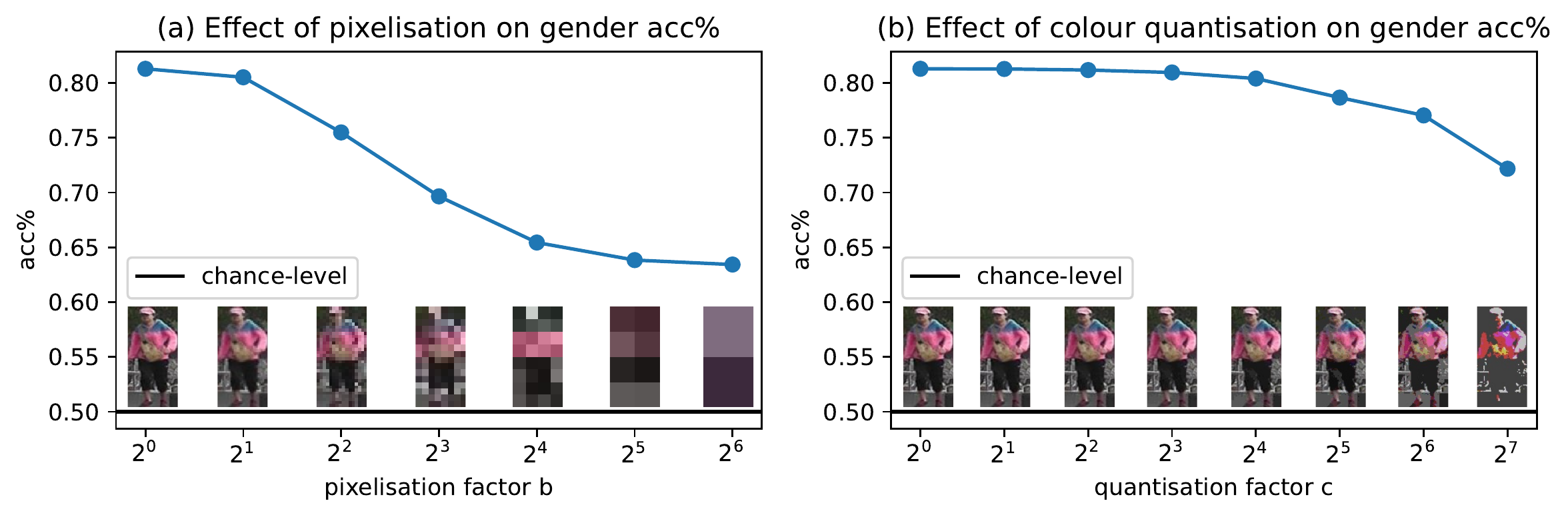}
\caption{Market1501 gender classification results, with varying $b$ and $c$.}
\label{fig:market_attr_dimreduc}
\end{figure*}

\figref{market_attr_dimreduc} shows the effect of increasingly large values for $b$ and $c$ on gender classification for Market1501 images. Unsurprisingly, the effect of pixelisation is more marked than that of colour quantisation, with the lowest accuracy being 63.4\% for the maximum $b$ value 64, against 72.2\% for the maximum $c$ value 128. While both are a significant departure from the initial classification accuracy of 81.3\%, neither suffice for achieving chance-level predictions.

Similarly, \tabref{fairface_attr_dimreduc} gives a summary of the effect of increasingly large dimensionality reduction parameters on gender, age, and ethnicity classification for FairFace images. Again, pixelisation has a larger impact than colour quantisation, and varying these parameters individually only reduces performances halfway to chance-level, at best. 

\begin{table}[h]
\centering
\caption{Fairface classification accuracy, with varying $b$ and $c$.}
\label{tab:fairface_attr_dimreduc}
\begin{tabular}{c|c|c|c}
\hline
\multirow{2}{*}{\textbf{Parameters}} & \multicolumn{3}{c}{\textbf{Accuracy\%}} \\
\cline{2-4}
& \textit{Gender} & \textit{Age} & \textit{Ethnicity} \\
\hline
\textit{b=1, c=1} & 79.8\% & 43.3\% & 46.1\% \\
\hline
\textit{b=2}$\mathit{^1}$\textit{, c=1} & 80.3\% & 43.9\% & 45.9\% \\
\textit{b=2}$\mathit{^3}$\textit{, c=1} & 76.9\% & 40.7\% & 43.4\% \\
\textit{b=2}$\mathit{^5}$\textit{, c=1} & 66.9\% & 32.2\% & 32.2\% \\
\hline
\textit{b=1, c=2}$\mathit{^1}$ & 80.2\% & 43.6\% & 46.1\% \\
\textit{b=1, c=2}$\mathit{^3}$ & 79.7\% & 43.1\% & 45.9\% \\
\textit{b=1, c=2}$\mathit{^5}$ & 76.4\% & 40.0\% & 41.4\% \\
\textit{b=1, c=2}$\mathit{^7}$ & 69.8\% & 33.6\% & 33.7\% \\
\hline
\textit{chance} & 50.0\% & 11.1\% & 14.3\% \\
\hline
\end{tabular}
\end{table}

\subsection{Limits of adverse task reduction}
Both our experiments and the observations described in the previous section showcase the ever-present tradeoff between utility and privacy; by design, increasing privacy, and thus limiting the information contained within data, comes at the cost of utility.

In our case, we are trying to conduct a task benefiting from high-dimensional data, i.e., person re-identification, while limiting the extent to which tasks such as demographic attribute classification can be carried out, which can do with much lower-dimensional data. As shown in \figref{market_attr_dimreduc}(a), using just two RGB pixels summarising the general colour information in images still allows for significantly above chance-level gender predictions; if any reID-relevant information is to be maintained within images, a utility-driven privacy mechanism can then hardly be expected to reduce demographic attribute classification performances beyond that level.

The ability of our utility-driven privacy mechanism to reduce gender prediction accuracy on Market1501 to $\sim$63.9\% (average of all experimented parameter \& noise combinations) is thus in line with the practical limit to adverse task performance reduction, which we observed to be 63.4\% with nothing but 2 RGB pixels of information, while retaining sensibly higher reID performances.

The same observation can be made regarding the Fairface dataset, where the application of our privacy mechanism reduces gender, age, and ethnicity prediction accuracies to averages of $\sim$66.0\%, $\sim$32.0\%, and $\sim$31.5\% (average of all experimented noise combinations), respectively, in line with the accuracies observed when applying exceedingly high pixelisation, which were 66.9\%, 32.2\%, and 32.2\%, respectively, for images composed of just 9 RGB pixels.

\section{Conclusion}
\label{sec:conclusion}
We introduce a strict pixel-level image differential privacy mechanism, which aims for indistinguishability between any pair of same-sized RGB images, leveraging both pixelisation and colour quantisation to bound the large sensitivity and therefore large noise this would otherwise entail. We show that by applying our privacy mechanism to an image dataset, one can sensibly reduce the utility of said data for very simple tasks such as gender, age, or ethnicity classification, to performance levels in line with those obtained by means of exceptionally high pixelisation-based obfuscation. Despite this low general-purpose utility, the images can still be used for near state-of-the-art person re-identification when using centroid-based models. We expect these results to be useful for building privacy-compliant camera-based pedestrian flow information systems, able to link together highly noised person representations without compromising pedestrians' privacy.

\bibliographystyle{IEEEtran}
\bibliography{references}

\begin{thebibliography}{10}
\providecommand{\url}[1]{#1}
\csname url@samestyle\endcsname
\providecommand{\newblock}{\relax}
\providecommand{\bibinfo}[2]{#2}
\providecommand{\BIBentrySTDinterwordspacing}{\spaceskip=0pt\relax}
\providecommand{\BIBentryALTinterwordstretchfactor}{4}
\providecommand{\BIBentryALTinterwordspacing}{\spaceskip=\fontdimen2\font plus
\BIBentryALTinterwordstretchfactor\fontdimen3\font minus
  \fontdimen4\font\relax}
\providecommand{\BIBforeignlanguage}[2]{{%
\expandafter\ifx\csname l@#1\endcsname\relax
\typeout{** WARNING: IEEEtran.bst: No hyphenation pattern has been}%
\typeout{** loaded for the language `#1'. Using the pattern for}%
\typeout{** the default language instead.}%
\else
\language=\csname l@#1\endcsname
\fi
#2}}
\providecommand{\BIBdecl}{\relax}
\BIBdecl

\bibitem{deakin2011smartcity}
\BIBentryALTinterwordspacing
M.~Deakin and H.~A. Waer, ``From intelligent to smart cities,''
  \emph{Intelligent Buildings International}, vol.~3, pp. 140--152, 2011.
  [Online]. Available: \url{https://doi.org/10.1080/17508975.2011.586673}
\BIBentrySTDinterwordspacing

\bibitem{schieder2021japan}
\BIBentryALTinterwordspacing
C.~S. Schieder, ``{Demographic decline: Reconciling population and social
  expectations},'' \emph{East Asia Forum Quarterly}, vol.~13, no.~3, pp.
  32--34, 2021. [Online]. Available:
  \url{https://search.informit.org/doi/10.3316/INFORMIT.114624371616764}
\BIBentrySTDinterwordspacing

\bibitem{schrotter2020urbanplanning}
\BIBentryALTinterwordspacing
G.~Schrotter and C.~H{\"u}rzeler, ``The digital twin of the city of zurich for
  urban planning,'' \emph{PFG--Journal of Photogrammetry, Remote Sensing and
  Geoinformation Science}, vol.~88, no.~1, pp. 99--112, 2020. [Online].
  Available: \url{https://doi.org/10.1007/s41064-020-00092-2}
\BIBentrySTDinterwordspacing

\bibitem{khan2021transport}
\BIBentryALTinterwordspacing
J.~Khan, R.~Hrelja, and F.~Pettersson-L^^c3^^b6fstedt, ``Increasing public
  transport patronage ^^e2^^80^^93 an analysis of planning principles and
  public transport governance in swedish regions with the highest growth in
  ridership,'' \emph{Case Studies on Transport Policy}, vol.~9, no.~1, pp.
  260--270, 2021. [Online]. Available:
  \url{https://doi.org/10.1016/j.cstp.2020.12.008}
\BIBentrySTDinterwordspacing

\bibitem{isoda2020routeplanning}
\BIBentryALTinterwordspacing
S.~Isoda, M.~Hidaka, Y.~Matsuda, H.~Suwa, and K.~Yasumoto, ``Timeliness-aware
  on-site planning method for tour navigation,'' \emph{Smart Cities}, vol.~3,
  no.~4, pp. 1383--1404, 2020. [Online]. Available:
  \url{https://doi.org/10.3390/smartcities3040066}
\BIBentrySTDinterwordspacing

\bibitem{ye2022reidreview}
\BIBentryALTinterwordspacing
M.~Ye, J.~Shen, G.~Lin, T.~Xiang, L.~Shao, and S.~H. Hoi, ``Deep learning for
  person re-identification: A survey and outlook,'' \emph{IEEE Transactions on
  Pattern Analysis \& Machine Intelligence}, vol.~44, no.~06, pp. 2872--2893,
  2022. [Online]. Available: \url{https://doi.org/10.1109/TPAMI.2021.3054775}
\BIBentrySTDinterwordspacing

\bibitem{bazzani2010appearance}
\BIBentryALTinterwordspacing
L.~Bazzani, M.~Cristani, A.~Perina, M.~Farenzena, and V.~Murino,
  ``Multiple-shot person re-identification by hpe signature,'' in \emph{20th
  International Conference on Pattern Recognition}, ser. ICPR'10, 2010, pp.
  1413--1416. [Online]. Available: \url{https://doi.org/10.1109/ICPR.2010.349}
\BIBentrySTDinterwordspacing

\bibitem{farenzena2010appearance}
\BIBentryALTinterwordspacing
M.~Farenzena, L.~Bazzani, A.~Perina, V.~Murino, and M.~Cristani, ``Person
  re-identification by symmetry-driven accumulation of local features,'' in
  \emph{2010 IEEE Computer Society Conference on Computer Vision and Pattern
  Recognition}, ser. CVPR'10, 2010, pp. 2360--2367. [Online]. Available:
  \url{https://doi.org/10.1109/CVPR.2010.5539926}
\BIBentrySTDinterwordspacing

\bibitem{krizhevsky2012cnn}
\BIBentryALTinterwordspacing
A.~Krizhevsky, I.~Sutskever, and G.~E. Hinton, ``Imagenet classification with
  deep convolutional neural networks,'' in \emph{Advances in Neural Information
  Processing Systems}, ser. NIPS'12, 2012, pp. 1097--1105. [Online]. Available:
  \url{https://dl.acm.org/doi/10.5555/2999134.2999257}
\BIBentrySTDinterwordspacing

\bibitem{yi2014deepreid}
\BIBentryALTinterwordspacing
D.~Yi, Z.~Lei, S.~Liao, and S.~Z. Li, ``Deep metric learning for person
  re-identification,'' in \emph{2014 22nd international conference on pattern
  recognition}, ser. ICPR'14, 2014, pp. 34--39. [Online]. Available:
  \url{https://doi.org/10.1109/ICPR.2014.16}
\BIBentrySTDinterwordspacing

\bibitem{zheng2015market1501}
\BIBentryALTinterwordspacing
L.~Zheng, L.~Shen, L.~Tian, S.~Wang, J.~Wang, and Q.~Tian, ``Scalable person
  re-identification: A benchmark,'' in \emph{2015 IEEE International Conference
  on Computer Vision}, ser. ICCV'15, 2015, pp. 1116--1124. [Online]. Available:
  \url{https://doi.org/10.1109/ICCV.2015.133}
\BIBentrySTDinterwordspacing

\bibitem{wei2017msmt17}
\BIBentryALTinterwordspacing
L.~Wei, S.~Zhang, W.~Gao, and Q.~Tian, ``Person transfer gan to bridge domain
  gap for person re-identification,'' in \emph{2018 IEEE/CVF Conference on
  Computer Vision and Pattern Recognition}, ser. CVPR'18, 2017, pp. 79--88.
  [Online]. Available: \url{https://doi.org/10.1109/CVPR.2018.00016}
\BIBentrySTDinterwordspacing

\bibitem{he2016resnet}
\BIBentryALTinterwordspacing
K.~He, X.~Zhang, S.~Ren, and J.~Sun, ``Deep residual learning for image
  recognition,'' in \emph{2016 IEEE Conference on Computer Vision and Pattern
  Recognition}, ser. CVPR'16, 2016, pp. 770--778. [Online]. Available:
  \url{https://doi.org/10.1109/CVPR.2016.90}
\BIBentrySTDinterwordspacing

\bibitem{hermans2017tripletloss}
\BIBentryALTinterwordspacing
A.~Hermans, L.~Beyer, and B.~Leibe, ``{In Defense of the Triplet Loss for
  Person Re-Identification},'' \emph{arXiv preprint arXiv:1703.07737}, 2017.
  [Online]. Available: \url{https://arxiv.org/abs/1703.07737}
\BIBentrySTDinterwordspacing

\bibitem{wang2018spatialtemporal}
\BIBentryALTinterwordspacing
G.~Wang, J.~Lai, P.~Huang, and X.~Xie, ``{Spatial-Temporal Person
  Re-identification},'' in \emph{Proceedings of the AAAI Conference on
  Artificial Intelligence}, vol.~33, no.~01, 2019, pp. 8933--8940. [Online].
  Available: \url{https://doi.org/10.1609/aaai.v33i01.33018933}
\BIBentrySTDinterwordspacing

\bibitem{fu2021unsupervised}
\BIBentryALTinterwordspacing
D.~Fu, D.~Chen, J.~Bao, H.~Yang, L.~Yuan, L.~Zhang, H.~Li, and D.~Chen,
  ``Unsupervised pre-training for person re-identification,'' in \emph{2021
  IEEE/CVF Conference on Computer Vision and Pattern Recognition}, ser.
  CVPR'21, 2021, pp. 14\,745--14\,754. [Online]. Available:
  \url{https://doi.org/10.1109/CVPR46437.2021.01451}
\BIBentrySTDinterwordspacing

\bibitem{chen2022attention}
\BIBentryALTinterwordspacing
Y.~Chen, H.~Wang, X.~Sun, B.~Fan, C.~Tang, and H.~Zeng, ``Deep attention aware
  feature learning for person re-identification,'' \emph{Pattern Recognition},
  vol. 126, p. 108567, 2022. [Online]. Available:
  \url{https://doi.org/10.1016/j.patcog.2022.108567}
\BIBentrySTDinterwordspacing

\bibitem{dwork2014dp}
\BIBentryALTinterwordspacing
C.~Dwork, A.~Roth \emph{et~al.}, ``The algorithmic foundations of differential
  privacy,'' \emph{Foundations and Trends{\textregistered} in Theoretical
  Computer Science}, vol.~9, no. 3--4, pp. 211--407, 2014. [Online]. Available:
  \url{https://doi.org/10.1561/0400000042}
\BIBentrySTDinterwordspacing

\bibitem{andres2013lgeodp}
\BIBentryALTinterwordspacing
M.~E. Andr\'{e}s, N.~E. Bordenabe, K.~Chatzikokolakis, and C.~Palamidessi,
  ``{Geo-Indistinguishability: Differential Privacy for Location-Based
  Systems},'' in \emph{Proceedings of the 2013 ACM SIGSAC Conference on
  Computer \& Communications Security}, ser. CCS'13, 2013, pp. 901--914.
  [Online]. Available: \url{https://doi.org/10.1145/2508859.2516735}
\BIBentrySTDinterwordspacing

\bibitem{abadi2016nndp}
\BIBentryALTinterwordspacing
M.~Abadi, A.~Chu, I.~Goodfellow, H.~B. McMahan, I.~Mironov, K.~Talwar, and
  L.~Zhang, ``Deep learning with differential privacy,'' in \emph{Proceedings
  of the 2016 ACM SIGSAC Conference on Computer and Communications Security},
  ser. CCS'16, 2016, pp. 308--318. [Online]. Available:
  \url{https://doi.org/10.1145/2976749.2978318}
\BIBentrySTDinterwordspacing

\bibitem{fan2018pixeldp}
\BIBentryALTinterwordspacing
L.~Fan, ``{Image Pixelization with Differential Privacy},'' in \emph{Data and
  Applications Security and Privacy XXXII}, vol. 10980, 2018, pp. 148--162.
  [Online]. Available: \url{https://doi.org/10.1007/978-3-319-95729-6_10}
\BIBentrySTDinterwordspacing

\bibitem{liu2021autoencoderdp}
\BIBentryALTinterwordspacing
B.~Liu, M.~Ding, H.~Xue, T.~Zhu, D.~Ye, L.~Song, and W.~Zhou, ``{DP-Image:
  Differential Privacy for Image Data in Feature Space},'' \emph{arXiv preprint
  arXiv:2103.07073}, 2021. [Online]. Available:
  \url{https://arxiv.org/abs/2103.07073}
\BIBentrySTDinterwordspacing

\bibitem{li2021gandp}
\BIBentryALTinterwordspacing
T.~Li and C.~Clifton, ``{Differentially Private Imaging via Latent Space
  Manipulation},'' \emph{arXiv preprint arXiv:2103.05472}, 2021. [Online].
  Available: \url{https://doi.org/10.48550/arXiv.2103.05472}
\BIBentrySTDinterwordspacing

\bibitem{zhao2022unstructureddp}
\BIBentryALTinterwordspacing
Y.~Zhao and J.~Chen, ``A survey on differential privacy for unstructured data
  content,'' \emph{ACM Computing Surveys}, vol.~54, no. 10s, pp. 1--28, 2022.
  [Online]. Available: \url{https://doi.org/10.1145/3490237}
\BIBentrySTDinterwordspacing

\bibitem{wieczorek2021centroids}
\BIBentryALTinterwordspacing
M.~Wieczorek, B.~Rychalska, and J.~Dabrowski, ``On the unreasonable
  effectiveness of centroids in image retrieval,'' in \emph{Neural Information
  Processing}, ser. ICONIP'21, vol. 13111, 2021, pp. 212--223. [Online].
  Available: \url{{https://doi.org/10.1007/978-3-030-92273-3_18}}
\BIBentrySTDinterwordspacing

\bibitem{russakovsky2015imagenet}
\BIBentryALTinterwordspacing
O.~Russakovsky, J.~Deng, H.~Su, J.~Krause, S.~Satheesh, S.~Ma, Z.~Huang,
  A.~Karpathy, A.~Khosla, M.~Bernstein, A.~C. Berg, and L.~Fei-Fei, ``{ImageNet
  Large Scale Visual Recognition Challenge},'' \emph{International Journal of
  Computer Vision (IJCV)}, vol. 115, no.~3, pp. 211--252, 2015. [Online].
  Available: \url{https://doi.org/10.1007/s11263-015-0816-y}
\BIBentrySTDinterwordspacing

\bibitem{lin2019marketattr}
\BIBentryALTinterwordspacing
Y.~Lin, L.~Zheng, Z.~Zheng, Y.~Wu, Z.~Hu, C.~Yan, and Y.~Yang, ``Improving
  person re-identification by attribute and identity learning,'' \emph{Pattern
  Recognition}, 2019. [Online]. Available:
  \url{https://doi.org/10.1016/j.patcog.2019.06.006}
\BIBentrySTDinterwordspacing

\bibitem{karkkainen2021fairface}
\BIBentryALTinterwordspacing
K.~Karkkainen and J.~Joo, ``{FairFace: Face Attribute Dataset for Balanced
  Race, Gender, and Age for Bias Measurement and Mitigation},'' in \emph{2021
  IEEE Winter Conference on Applications of Computer Vision}, ser. WACV'21,
  2021, pp. 1547--1557. [Online]. Available:
  \url{https://doi.org/10.1109/WACV48630.2021.00159}
\BIBentrySTDinterwordspacing

\end{thebibliography}

\end{document}